\documentclass[10pt,twocolumn,letterpaper]{article}

\usepackage[pagenumbers]{cvpr} 

\usepackage{graphicx}
\usepackage{amsmath}
\usepackage{amssymb}
\usepackage{booktabs}
\usepackage{enumitem}
\DeclareMathOperator*{\argmax}{arg\,max}

\usepackage{multirow}
\usepackage{amsmath}
\usepackage[accsupp]{axessibility}

\usepackage[pagebackref,breaklinks,colorlinks]{hyperref}

\usepackage[capitalize]{cleveref}
\crefname{section}{Sec.}{Secs.}
\Crefname{section}{Section}{Sections}
\Crefname{table}{Table}{Tables}
\crefname{table}{Tab.}{Tabs.}

\def\cvprPaperID{7027} 
\def\confName{CVPR}
\def\confYear{2023}

\begin{document}

\title{TopNet: Transformer-based Object Placement Network for Image Compositing}

\author{Sijie Zhu$^{1}$, Zhe Lin$^{2}$, Scott Cohen$^{2}$, Jason Kuen$^{2}$, Zhifei Zhang$^{2}$, Chen Chen$^{1}$ \\
$^{1}$Center for Research in Computer Vision, University of Central Florida \quad $^{2}$Adobe Research \\
{\tt\small sizhu@knights.ucf.edu,\{zlin,scohen,kuen,zzhang\}@adobe.com,chen.chen@crcv.ucf.edu}}

\maketitle

\begin{abstract}
   We investigate the problem of automatically placing an object into a background image for image compositing. Given a background image and a segmented object, the goal is to train a model to predict plausible placements (location and scale) of the object for compositing. The quality of the composite image highly depends on the predicted location/scale. Existing works either generate candidate bounding boxes or apply sliding-window search using global representations from background and object images, which fail to model local information in background images. However, local clues in background images are important to determine the compatibility of placing the objects with certain locations/scales. In this paper, we propose to learn the correlation between object features and all local background features with a transformer module so that detailed information can be provided on all possible location/scale configurations. A sparse contrastive loss is further proposed to train our model with sparse supervision. Our new formulation generates a 3D heatmap indicating the plausibility of all location/scale combinations in one network forward pass, which is $>10\times$ faster than the previous sliding-window method. It also supports interactive search when users provide a pre-defined location or scale. The proposed method can be trained with explicit annotation or in a self-supervised manner using an off-the-shelf inpainting model, and it outperforms state-of-the-art methods significantly. The user study shows that the trained model generalizes well to real-world images with diverse challenging scenes and object categories. 
\end{abstract}

\section{Introduction}
\label{sec:intro}
Object compositing~\cite{niu2021making,zhang2021deep} is a common and important workflow for image editing and creation. The goal is to insert an object from an image into a given background image such that the resulting image appears visually pleasing and realistic. Conventional workflows in object compositing rely on manual object placement, \ie manually determining where the object should be placed (location) and in what size the object is placed (scale). However, the manual placement does not fulfill the growing need for image creation for social sharing, advertising, education, etc., and AI-assisted compositing with automatic object placement is more desirable for future image creation applications. While there have been several works on learning-based object placement for specific scenes, general object placement with diverse scenes and objects still remains challenging with limited exploration, as it involves a deeper understanding of common sense, objects, and local details of scenes. Inaccurate object placement could lead to poor compositing results, \eg a person floating in the sky, a dog larger than buildings, etc. 
\begin{figure*}[!htbp]
    \centering
    \includegraphics[width=.9\linewidth]{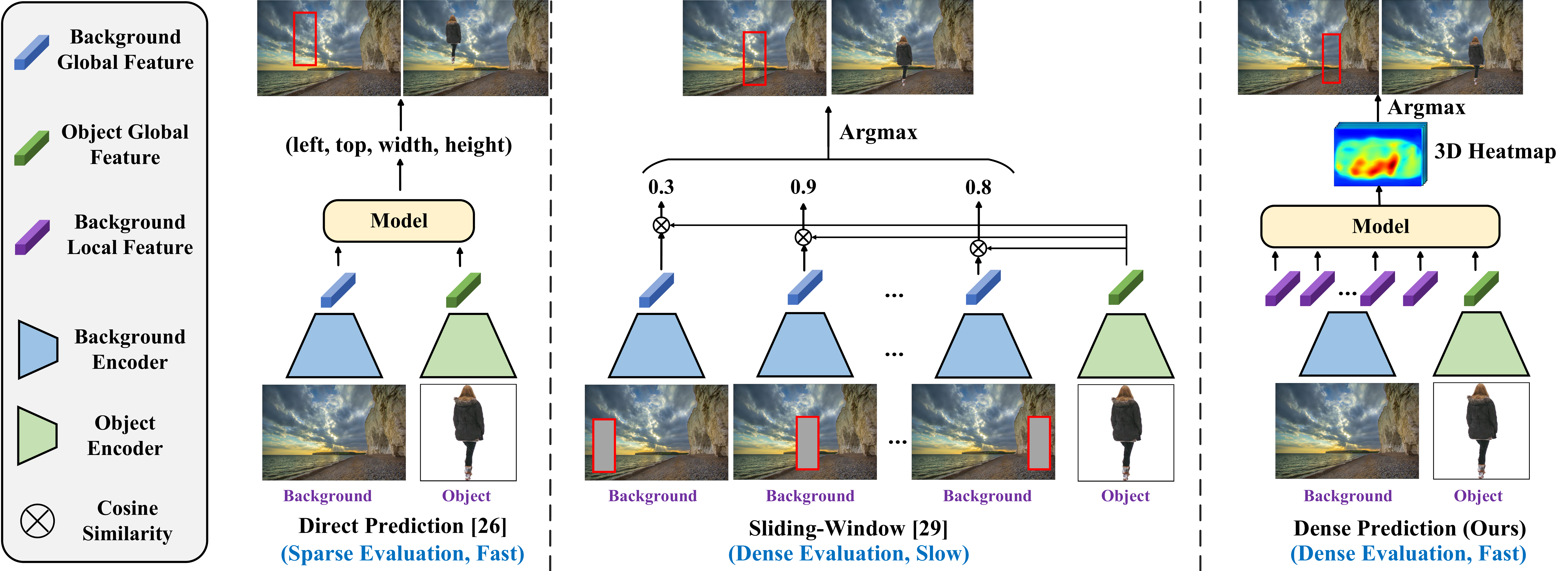}
    \vspace{-0.2cm}
    \caption{Comparison between different formulations for object placement. Our method provides a dense evaluation of possible locations/scales in one network forward pass.}
    \label{fig:compare}
    \vspace{-0.1cm}
\end{figure*}

Existing works \cite{lee2018context,lin2018st, zhang2020learning,liu2021OPA,zhu2022gala} formulate the problem in very different ways, as shown in Fig.~\ref{fig:compare}. \cite{lin2018st,zhang2020learning} directly predict multiple transformations or bounding boxes indicating the location and scale of the given objects. Such sparse predictions recommend the top candidate placements for users, but they do not provide any information about other possible locations and scales. They also fail to leverage the local clues in background images, as the bounding boxes are generated based on only global features. Another thread of works \cite{ECCV_workshop,liu2021OPA} considers object placement as binary classification, which evaluates the plausibility of input images and placement of bounding boxes instead of generating candidate placements directly from input images. One recent work \cite{zhu2022gala} utilizes a retrieval model to assess the plausibility of a given placement and evaluates a grid of locations and scales in a sliding-window manner. However, it requires multiple network forward passes to generate a dense evaluation for one image, resulting in a slow inference speed. 

In this paper, we propose TopNet, a Transformer-based Object Placement Network for real-world object compositing applications. Different from previous works, TopNet formulates object placement as a dense prediction problem: \textit{generating evaluation for a dense grid of locations and scales in one network forward pass.} Given a background image and an object, TopNet directly generates a 3D heatmap indicating the plausibility score of object location and scale, which is $>10\times$ faster than the previous sliding-window method \cite{zhu2022gala}. Previous works \cite{zhang2020learning,zhu2022gala} combine background and foreground object features only at the global level, which fails to capture local clues for determining the object location.
We propose to learn the correlation between global foreground object features and local background features with a multi-layer transformer, leading to a more efficient and accurate evaluation of all possible placements. To train TopNet with sparse supervision where only one ground-truth placement bounding box is provided, we propose a sparse contrastive loss to encourage the ground-truth location/scale combination to have a relatively high score, while only minimizing the other combinations with the lowest score or a score higher than the ground-truth with a certain margin, thus preventing large penalty on other reasonable locations/scales. Once the 3D heatmap is predicted, top candidate placement bounding boxes can be generated by searching the local maximum in the 3D heatmap. The 3D heatmap also provides guidance for other possible locations/scales which are not the best candidate. 
Experiments on a large-scale inpainted dataset (Pixabay \cite{pixabay}) and annotated dataset (OPA \cite{liu2021OPA}) show the superiority of our approach over previous methods. Our contributions are summarized as follows:
\setlist{nolistsep}
\begin{itemize}[noitemsep,leftmargin=*]
    \item A novel transformer-based architecture to model the correlation between object image and local clues from the background image, and generate dense object placement evaluation $>10\times$ faster than previous sliding-window method \cite{zhu2022gala}.
    \item A sparse contrastive loss to effectively train a dense prediction network with sparse supervision.
    \item Extensive experiments on a large-scale inpainted dataset and annotated dataset with state-of-the-art performance.
\end{itemize}
\begin{figure*}
    \centering
    \includegraphics[width=.8\linewidth]{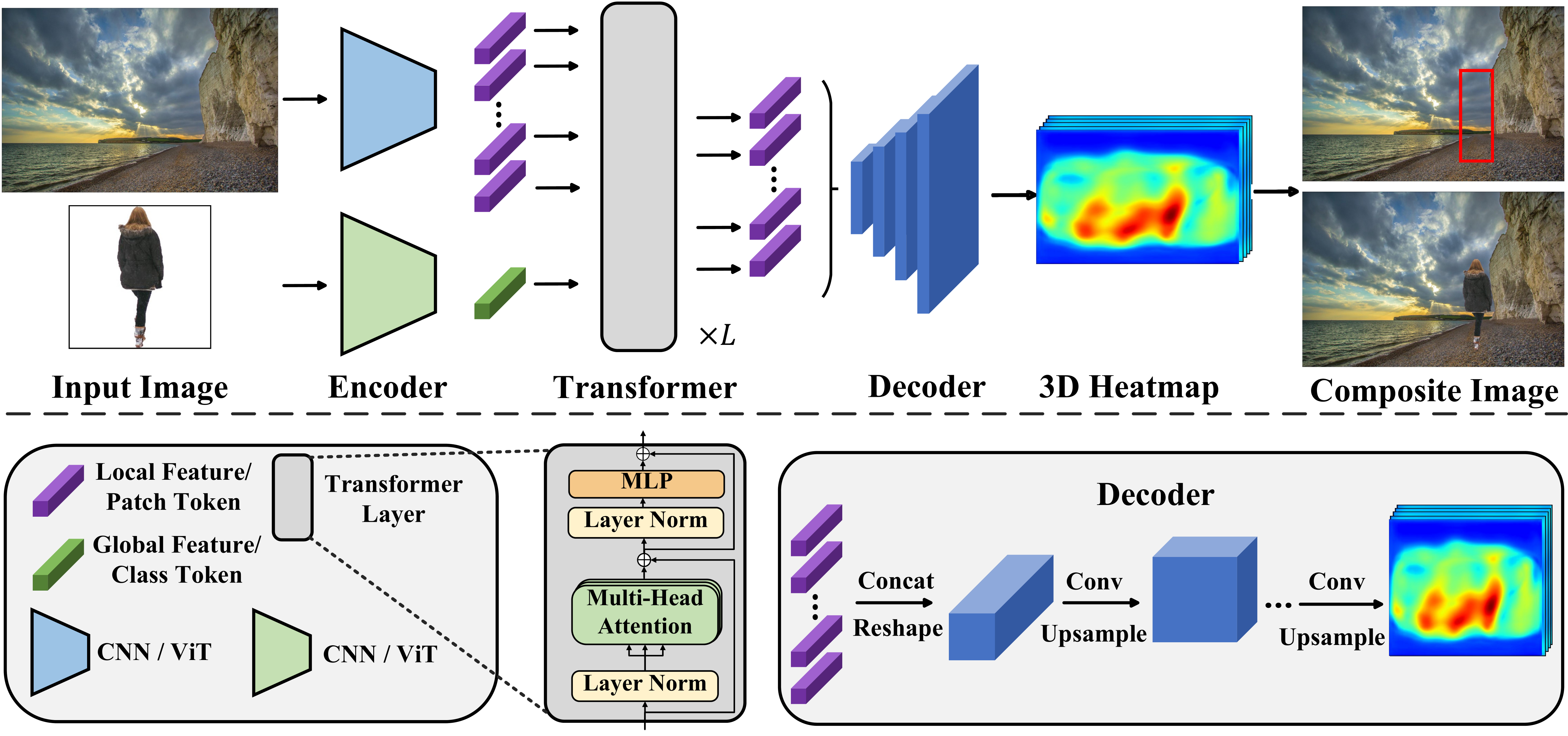}
    \vspace{-0.2cm}
    \caption{An overview of the proposed TopNet. Local background features and global object features are extracted with two different encoders and fed into a transformer module to learn the correlation. A CNN-based upsampling decoder is then adopted to generate a 3D heatmap. Top-1/top-5 bounding boxes can be generated by finding the global/local minimum in the 3D heatmap.}
    \label{fig:architecture}
    \vspace{-0.1cm}
\end{figure*}
\section{Related Work}
\label{sec:related}
\textbf{Object Placement Prediction.}
There are several works learning to directly predict object placement as a bounding box or transformation. Tan \etal \cite{tan2018and} propose to predict bounding boxes for person objects using background and layout images. Then retrieval is applied to find a specific person object for compositing. ST-GAN \cite{lin2018st} models the compositing realism in geometric wrap parameter space and learns to predict geometric transformation using adversary training. 
Similarly, Tripathi \etal \cite{tripathi2019learning} train a synthesizer network to predict transformation along with a discriminator to tell if the composite image is real. CompositionalGAN \cite{azadi2020compositional} designs a self-consistent network for compositing so that composite images can be decomposed back into individual objects. Li \etal \cite{li2019putting} focus on indoor scenes by simultaneously learning location and plausible human poses. Lee \etal \cite{lee2018context} propose to predict object masks for certain categories and then determine suitable object instances to be inserted. PlaceNet \cite{zhang2020learning} is close to our setting as it can also be trained on inpainted background images with the original foreground object. It predicts bounding boxes with global features and random inputs and trains a discriminator to determine the plausibility of the bounding boxes along with global features of background and object images. These works either deal with street scenes with limited object categories (\eg persons, vehicles, traffic lights) or indoor scenes with human pose joints or furniture. \emph{We focus on real-world object placement for general compositing with diverse scenes and object categories, which is more challenging and not well studied in this field.}\\
\textbf{Compositing Evaluation.}
Several recent works focus on object placement evaluation for general compositing with diverse scenes and object categories. OPA \cite{liu2021OPA} proposes to train a binary classifier to determine whether a certain placement is realistic using both the composite image and placement mask. It is trained with human annotation on a subset of COCO dataset. GALA \cite{zhu2022gala} trains a retrieval network to find the best object given a background image with placement bounding box, which could be considered as a single evaluation on certain placement. It is trained with contrastive learning on the masked background and the corresponding foreground objects without human annotation. GALA further demonstrates that such a single evaluation can be extended with a sliding-window grid search to predict object placement. Although it is annotation-free and performs well on small-scale datasets, its inference time is proportional to the search space which could be large in practical scenarios. \emph{We tackle this problem efficiently by generating dense evaluation in one network forward pass.}\\
\textbf{Object Detection/Segmentation.} Objection detection \cite{tian2019fcos} and segmentation \cite{ronneberger2015u,he2017mask,lee2020centermask} are related because they also generate dense prediction with bounding boxes or heatmaps.  Recent segmentation methods \cite{zheng2021rethinking,lee2020centermask} generally follow the design of an encoder-decoder framework, where the encoder extracts high-level features from images and the decoder applies convolutional operations and upsampling to generate a segmentation map with high resolution. \emph{However, these works cannot be directly applied to our problem.} First, detection and segmentation take one background image as input, while our task has another object image as input. Second, detection and segmentation have dense supervision, \ie ground-truth bounding boxes for all objects or pixel-level segmentation masks, but our task usually has only one ground-truth bounding box as sparse supervision. \emph{The proposed novel architecture and loss function are specifically designed to tackle these two issues.}  

\section{Methodology}
\label{sec:method}
\subsection{Formulation}
\label{sec:formulation}
Given a background image $I_{b}\in \mathbb{R}^{h_{b}\times w_{b} \times 3}$ and a foreground object image $I_{o} \in \mathbb{R}^{h_{o} \times w_{o} \times 3}$, typical object placement is represented as a bounding box $[l, t, w, h]$, indicating the left, top, width, and height of the object in final composite image $I_{c}\in \mathbb{R}^{h_{b}\times w_{b} \times 3}$. Since the aspect ratio of the object image is known, we assume that the aspect ratio of the object is kept as $w_{o}/h_{o}$ so that the scale can be defined as one number $s=\sqrt{wh/w_{b}h_{b}}$. Then the dense evaluation model predicts a 3D heatmap $H\in \mathbb{R}^{h_{b}\times w_{b}\times c}$, where $c$ is the number of pre-defined scales, indicating the plausibility of each location-scale combination. Each spatial location in the heatmap corresponds to the center of a placement bounding box. By default, we use $c=16$ and each channel represents a scale value from $0.15$ to $0.9$ with an interval of $0.05$, according to the distribution of  detectable objects in their original background images. 

During inference, we first normalize $H$ with maximum and minimum value as $\hat{H}=(H-min(H))/(max(H)-min(H))$. The top-1 prediction can be easily generated as $\argmax(\hat{H})$, the global maximum location in the 3D heatmap. Top-k candidate predictions can be generated by finding local maximum peaks where the heatmap value is larger than a threshold. We use double standard deviation over average value as the threshold in the experiments.
\begin{figure*}
    \centering
    \includegraphics[width=.8\linewidth]{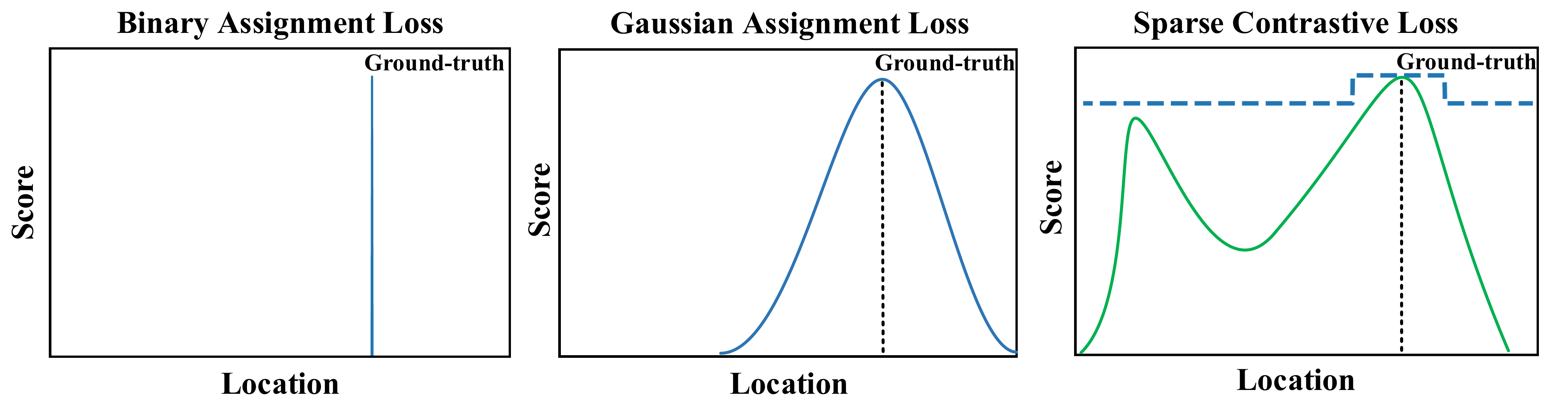}
    \vspace{-0.3cm}
    \caption{An example of different losses in 1D space. The solid blue curve denotes actual assignment, while the dash blue curve denotes upper bound assignment. The green solid curve illustrates one possible multi-peak score distribution, which is not allowed by the other assignment losses.}
    \label{fig:loss}
    \vspace{-0.2cm}
\end{figure*}

\subsection{Architecture}
\label{sec:architecture}

\textbf{Global and Local Features.} 
As shown in Fig. \ref{fig:architecture}, we adopt two encoder networks to learn different features for background and object images. The encoder could be CNN \cite{he2016deep} (Convolutional Neural Network) or ViT \cite{vit} (Vision Transformer). To determine whether a specific location is suitable for the object with a certain scale, the local clues in background images could provide detailed information. We thus keep all the local features/tokens from the last convolutional or transformer layer of the background encoder. For foreground objects, the image is relatively simple, so we only keep the global feature (\ie the last feature before the classification head), which encodes high-level information. \\
\textbf{Background and Object Correlation.} 
To learn the correlation between the background local features and global features of the object image, we adopt a multi-layer transformer \cite{transformer} module, which has been shown to model strong global attention well in vision tasks \cite{vit,deit}. We treat the background features as patch tokens \cite{vit} in ViT and add 2D version of sinusoidal positional embedding \cite{transformer} accordingly. The class token is then replaced with the global feature of the object image. We take all the patch tokens of the last layer as output to feed into the upsampling decoder. A $1\times 1$ convolutional layer is applied to the input features to convert the feature dimension to the dimension of transformer module. The inner architecture of a transformer layer is shown in Fig. \ref{fig:architecture}, and the key component is the multi-head self-attention module. In a basic self-attention module, each input token is first converted into query, key, and value, denoted as $Q, K, V$ with dimension $d$, with three learnable linear projections. Then attention map is computed as $softmax(QK^{T}/d)V$. A multi-head attention module runs multiple basic self-attention modules in parallel and concatenates the outputs from all the heads.\\
\textbf{Upsampling Decoder.} Given the input background image with size of $h_{b}\times w_{b} \times 3$, there are typically $\frac{h_{b}}{2^{k}}\frac{w_{b}}{2^{k}}$ local features with dimension $d$, where $2^{k}$ is the downsampling ratio. As shown in Fig. \ref{fig:architecture}, they are first concatenated and reshaped into size of $\frac{h_{b}}{2^{k}} \times \frac{w_{b}}{2^{k}} \times d$. Similar to the pyramid architecture in image segmentation \cite{zheng2021rethinking}, the decoder then applies $k$ convolutional layers where each one is followed by a $2\times$ spatial upsampling block. Each convolution layer has a kernel of $3\times 3$ and reduces the dimension by $2\times$. The last layer has an output dimension of $c$, which is the number of pre-defined scales. We set $k$ as $4$ in our experiments. 

\subsection{Loss Function}
\label{sec:loss}
The main challenge for dense location and scale evaluation is the sparse supervision signal. Small-scale annotation in CAIS~\cite{cais} or OPA~\cite{liu2021OPA} only provides one positive placement bounding box for each sample, without supervision on other locations/scales. For large-scale datasets without explicit annotation, one way to generate supervision is to mask out the original objects in background images and generate pure background images using off-the-shelf inpainting models. Then bounding box of the original object can be considered as the ground-truth placement, but the supervision is still sparse -- only one location and scale. 
\begin{figure*}[!htbp]
    \centering
    \includegraphics[width=.95\linewidth]{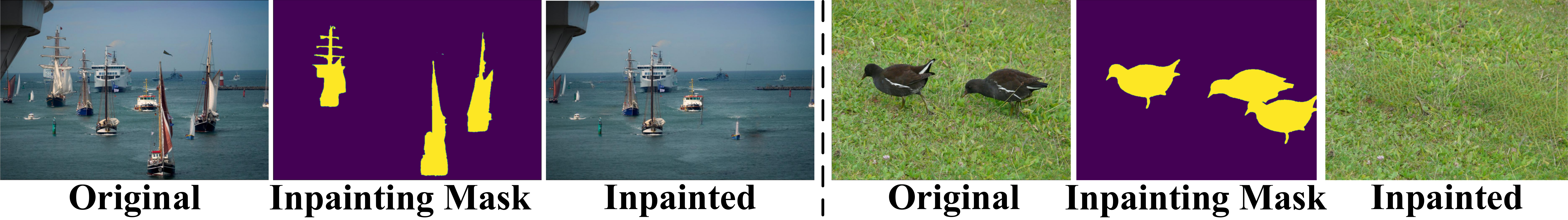}
    \vspace{-0.2cm}
    \caption{Example inpainted background images. If there are less than $3$ objects, we add one additional randomly shifted object mask to prevent the model from learning inpainting artifacts.}
    \label{fig:inpainting}
    \vspace{-0.2cm}
\end{figure*}

One simple idea to supervise the model is to assign a ground-truth score for each location-scale combination, \ie each data point in the 3D heatmap. Simple binary assignment considers the only ground-truth combination (GT data point) as $1$, and all other location-scale combinations as $0$. A smoother assignment is Gaussian assignment, which gives the score according to the distance between each data point and the ground-truth point in the 3D space. It considers locality of the score, \ie locations/scales close to the ground-truth should still be good placement candidates. These assignments consider all locations/scales far away from the ground-truth as negative points with low scores. However, such an assumption does not hold in most cases. Given a specific background scene, certain objects could be compatible at many locations with different scales. One could drag the object in Fig. \ref{fig:inpainting} along the ground toward the camera direction and increase the scale accordingly. However, considering local clues like light and aesthetic judgment, some location-scale combinations are still better than others. We thus assume that multiple good candidate bounding boxes exist in a background image, and propose to maximize the score at the ground-truth location/scale while allowing local peaks with high scores in other locations/scales. One toy example of different losses in 1D space is shown in Fig. \ref{fig:loss}. Assume that the ground-truth coordinate in the 3D heatmap $H \in \mathbb{R}^{h_{b}\times w_{b}\times c}$ is $(x_{gt},y_{gt},z_{gt})$. The first loss term is formulated as:
\begin{equation}
\small
\resizebox{0.91\linewidth}{!}{
    $\mathcal{L}_{con} = \sum_{(x,y,z)}|H(x,y,z) - H(x_{gt},y_{gt},z_{gt}) + M(x,y,z)|^{+},$}
\end{equation}
where $(x,y,z) \in \mathbb{R}^{h_{b}\times w_{b}\times c}$ and $|\cdot|^{+}$ means $max(\cdot,0)$. $M\in \mathbb{R}^{h_{b}\times w_{b}\times c}$ is the margin matrix, indicating how much $H(x_{gt},y_{gt},z_{gt})$ should be higher than $H(x,y,z)$ for any $(x,y,z)$. The margin is set as 0 for the neighborhood of ground-truth location/scale if $|x-x_{gt}|\le 20,|y-y_{gt}|\le 20,|z-z_{gt}|\le 2$. Otherwise, it is empirically set as $0.1$. We then apply the second term as: 
\begin{equation}
\small
\mathcal{L}_{range}=|1-H(x_{gt},y_{gt},z_{gt})|+|min(H)|,    
\end{equation}
so that the scores fall in $[0,1]$. It also encourages the lowest score to be $0$, as there is always a bad location or scale for certain background and object images. This term prevents the model from predicting a high score for all locations/scales. The overall spare contrastive loss is defined as the summation of two terms $\mathcal{L} = \mathcal{L}_{con}+\mathcal{L}_{range}$.

\section{Experiment}
\label{sec:experiment}
\subsection{Datasets and Implementation Detail}
\noindent\textbf{Pixabay} \cite{pixabay} is a dataset collected from ``pixabay.com'', a free stock image site. It contains millions of high-quality, diverse, free-to-use photos which are perfect for building and evaluating models for real-world object compositing. We follow \cite{zhu2022gala} to collect $928,018$ images and apply object detection \cite{tian2019fcos} and segmentation \cite{lee2020centermask} to generate foreground objects and background images. This results in $5,771,912$ objects and $928,018$ background images. We then filter out tiny objects that are unlikely to be used for composting, and background images with overly large object mask where no background information is left. We keep objects with high confidence detection scores and proper bounding box sizes, resulting in $833,964$ foreground and background pairs with $914$ non-zero categories. Finally, they are randomly split into training/evaluation sets with $90\%/10\%$ ratio.

Since there is no annotation for this large-scale dataset, we generate pure background images by removing objects from each image. The original bounding box of the object would always be a good placement candidate and we consider it as ground-truth. We adopt the off-the-shelf model of LAMA \cite{suvorov2022resolution} for inpainting. For images with three or more objects, we randomly select three object masks for inpainting. If there are fewer than three objects, we use all the object masks and add an additional mask by randomly shifting one of the object masks. In this way, multiple locations contain inpainting artifacts, which reduces the risk of model overfitting to inpainting artifacts. As shown in Fig. \ref{fig:inpainting}, the inpainting model \cite{suvorov2022resolution} works well on recovering the background sea and grass. Furthermore, we observe that images with too many objects have low inpainting quality due to strong occlusion. Therefore, we remove the images with more than $5$ objects, resulting in $367,384$ pairs for training and $41,166$ pairs for evaluation. Each pair contain one inpainted background and the original foreground object. \\
\textbf{OPA} \cite{liu2021OPA} dataset is proposed for object placement evaluation with human annotation for each composite image. In total, it contains $62,074$ training images and $11,396$ test images without overlap. All the images are collected from COCO \cite{lin2014microsoft} dataset and each composite image is generated with one background, one object, and one placement bounding box. We adopt OPA for our object placement prediction experiment, using only the positive samples. We consider the positive bounding box as ground-truth, resulting in $21,350$ image pairs for training and $3,566$ pairs for testing.\\
\textbf{Implementation Detail.} Our method is implemented based on PyTorch \cite{paszke2019pytorch} and trained on one RTX A5000 GPU. The batch size is set to 64 for all methods. By default, we use ViT-small \cite{vit} pre-trained weights \cite{deit} on ImageNet \cite{deng2009imagenet} as encoder backbone.
Object images are first padded with white pixels as square images before being fed into the encoder. Then both the object and background images are resized to $224\times 224$ and normalized with RGB average values. We adopt AdamW \cite{adamw} optimizer with a learning rate of 0.00001 and weight decay of $0.03$. The learning rate is adjusted with cosine scheduling \cite{loshchilov2016sgdr}.

\begin{table*}[!htbp]
\centering
\begin{tabular}{l c c c c c c}
\hline

\hline
\multirow{2}{*}{Method} & \multirow{2}{*}{Infer. Time (s)} & \multicolumn{2}{c}{\textbf{Pixabay}} & & \multicolumn{2}{c}{\textbf{OPA}} \\
\cline{3-4} \cline{6-7}
& & $IOU>0.5$ & Mean IOU & & $IOU>0.5$ & Mean IOU\\
\hline

\hline
Regression \cite{zhang2020learning} & 0.08 & 48.23 &  0.448  & & ~7.24 & 0.178 \\ 
$\dagger$Retrieval \cite{zhu2022gala} &  1.69 & 11.91 & 0.220 & & ~2.08 & 0.112\\ 
Classifier \cite{liu2021OPA} &  0.55 &  ~6.82 & 0.147 & & ~2.54 & 0.115 \\ 
PlaceNet \cite{zhang2020learning} &  0.16 &  19.44 & 0.308 & & 10.09 & 0.225\\ 
\hline
Ours  &  0.11  & \textbf{74.74} & \textbf{0.620} & & \textbf{15.95} & \textbf{0.241}\\ 
\hline

\hline
\end{tabular}
\caption{Evaluation on top-5 predictions in terms of maximum IOU between the top-5  predicted bounding boxes and the ground-truth.}
\label{tab:main}
\vspace{-0.2cm}
\end{table*}

\subsection{Evaluation Metric}
\label{sec:metric}
\noindent\textbf{Top-k IOU}. Given ground-truth bounding box and the predicted bounding boxes, one simple way to evaluate is to compute the IOU (Intersection over Union) between the ground-truth and top-1 predicted box. However, such evaluation tends to have $0$ IOU for lots of samples, as there are usually multiple good candidate locations and the top-1 prediction might be reasonable but not exactly at the ground-truth location. In practice, it would also be better to provide multiple candidates for users to select. We thus apply top-5 IOU as an evaluation metric, \ie the best IOU between ground-truth and top-5 predicted bounding boxes.  \\
\textbf{Normalized Score.} For methods with heatmap scores, the ground-truth may not be necessary to have the highest score, but it should be among the best ones. We thus apply the normalized score as one of the evaluation metrics. The heatmap score is first normalized with the minimum and maximum value as $\hat{H}$ in Sec. \ref{sec:formulation}, which is referred to as Normalized Score (NS). NS at the ground-truth location/scale may not be $1$ but should be relatively high as compared with other locations and scales. Therefore, we compute the mean NS and portion of NS above a certain threshold (\eg 0.9). The NS is more reasonable than IOU when only location is evaluated, as a small spatial shift could lead to a $0$ IOU.  

\begin{table}[tbp]
\small
\centering
\resizebox{\linewidth}{!}{
\begin{tabular}{l c c c c c}
\hline

\hline
\multirow{2}{*}{Method} & Infer. & \multicolumn{3}{c}{$NS$} & Mean \\
\cline{3-5}
&  Time (s) & $>0.95$ & $>0.9$ & $>0.75$ & $NS$($\uparrow$) \\ 
\hline

\hline
$\dagger$Retrieval \cite{zhu2022gala} &  1.35 &  8.00 & 20.32 & 57.84 & 0.75
\\ 
Classifier \cite{liu2021OPA} &  0.41 &  21.21 & 32.15 & 53.71 & 0.70
\\ 
\hline
Ours &  \textbf{0.11} &  \textbf{47.53} & \textbf{67.70} & \textbf{90.30} & \textbf{0.90} \\
\hline

\hline
\end{tabular}}
\caption{Evaluation on location prediction given ground-truth scale on Pixabay. NS (normalized score) denotes the predicted score at the ground truth location normalized by the maximum and minimum values across all locations.}
\label{tab:location}
\vspace{-0.1cm}
\end{table}

\subsection{Comparison with State-of-the-art}
We compare the proposed method with state-of-the-art and several baseline methods:
1) ``Regression'' \cite{zhang2020learning} simply trains the network to predict the ground-truth bounding box with MSE (Mean Square Error) loss. The regression head (MLP) is adopted on the concatenated global features of background and object images. 2) ``$\dagger$Retrieval'' follows the pipeline in \cite{zhu2022gala}, except we modify the initial bounding box size as the average size of our new dataset. $\dagger$ denotes that the retrieval model has a different setting than other methods. For Pixabay, the model is directly obtained from \cite{zhu2022gala} which was trained on a different version. It does not apply inpainting and has about $2\times$ images as compared with our filtered version. The original retrieval model is trained with the masked background and the corresponding foreground objects, thus does not require inpainting on background images. However, the OPA dataset does not provide such data. Therefore, we use the ground-truth object for training instead of the original object. 3) ``Classifier'' follows \cite{liu2021OPA} to train a binary classifier and predicts whether a composite image is reasonable. It is further extended with the sliding-window method in \cite{zhu2022gala} to generate location/scale, \ie generating composite images and masks with a grid of location/scale combinations and selecting the one with the maximum score. 4) PlaceNet \cite{zhang2020learning} follows the implementation in \cite{zhang2020learning} with adversary training. The bounding box is predicted based on the global features of background and object images along with a random vector, and a discriminator is trained to tell whether the bounding box is reasonable, conditioned on the global features.

In Table \ref{tab:main}, we show the top-5 IOU (Sec. \ref{sec:metric}) based evaluation on both Pixabay and OPA datasets. Top-1 IOU is included in \textbf{supplementary material}. The proposed method achieves significant improvements over state-of-the-art methods on both inpainted and manually annotated datasets, indicating the superiority of the dense prediction as compared with sparse prediction or sliding-window formulation. Remarkably, the inference speed of the proposed method is on-par with the single-step prediction (``Regression'') and is over \textbf{$>10\times$ faster} than the previous sliding-window based method, \ie ``$\dagger$Retrieval''. 

To show the effectiveness of location and scale prediction separately, we conduct experiments on interactive search where the users have provided the ground-truth location or scale. In our 3D heatmap, we simply set the pre-defined dimensions as the desired numbers and search for the other dimensions. For sliding-window-based methods (``$\dagger$Retrieval'', ``Classifier''), the sliding windows are applied only on the dimensions to be searched. For location prediction, we generate a 2D heatmap using all the methods and evaluate using the NS (Sec. \ref{sec:metric}). The sparse prediction methods like ``Regression'' are not feasible in this case. For scale prediction, we fix the location as the ground-truth location and compute the IOU between the ground-truth and prediction. The prediction of sparse prediction methods would not change in this case, while other methods can be improved using the additional information of ground-truth location. Tables \ref{tab:location} and \ref{tab:scale} show that the proposed method still significantly outperforms the previous methods.

\begin{figure*}[!htbp]
    \centering
    \includegraphics[width=.8\linewidth]{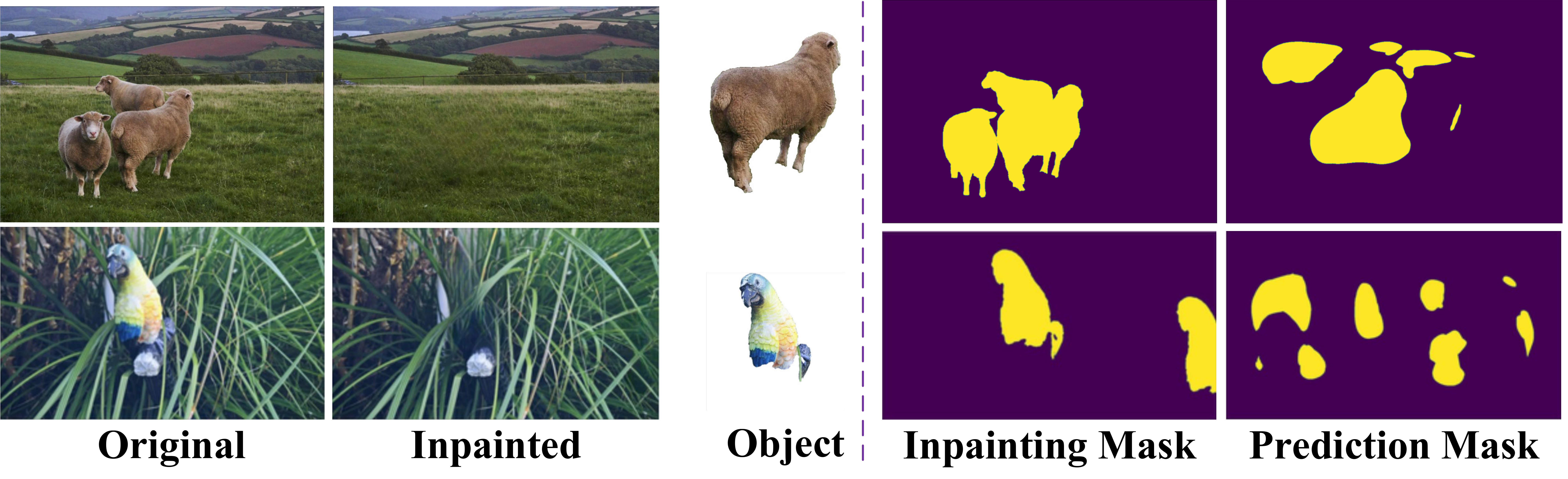}
    \vspace{-6pt}
    \caption{Example of inpainting mask and prediction mask. They usually do not have large overlapping regions.}
    \label{fig:artifact}
    \vspace{-0.1cm}
\end{figure*}

\begin{figure*}[!htbp]
    \centering
    \includegraphics[width=1.\linewidth]{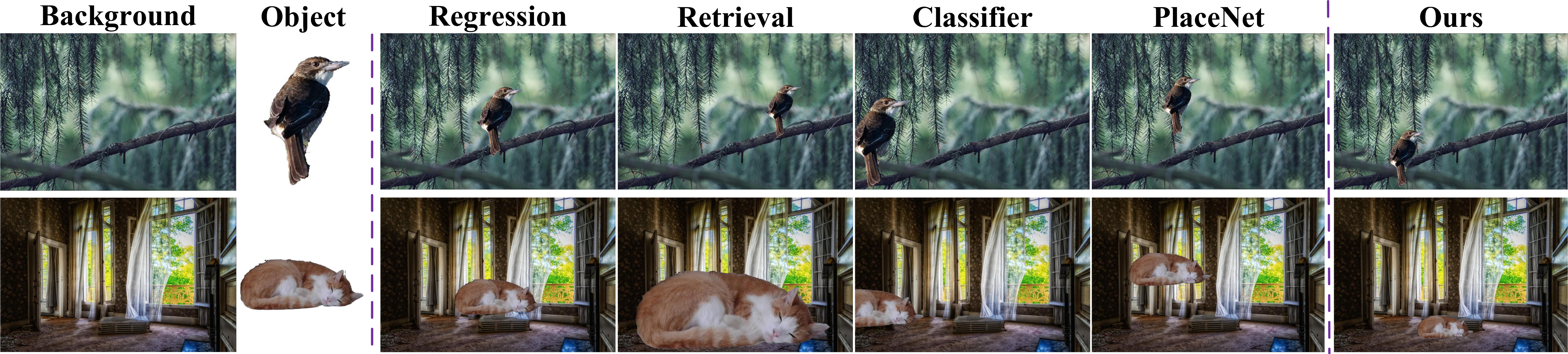}
    \vspace{-0.5cm}
    \caption{Qualitative results on real-world images. Best viewed with zoom-in. Tree branch and indoor room scenarios are very challenging.}
    \label{fig:user}
\end{figure*}

\begin{table}[tbp]
\small
\centering
\resizebox{\linewidth}{!}{
\begin{tabular}{l c c c c c}
\hline

\hline
\multirow{2}{*}{Method} & Infer. & \multicolumn{3}{c}{$IOU$} & Mean \\
\cline{3-5}
& Time (s) & $>0.95$ & $>0.9$ & $>0.75$ & Error($\downarrow$) \\
\hline

\hline
$\dagger$Retrieval \cite{zhu2022gala} &  0.32 & 15.71 & 31.50 & 68.05 & 0.105 \\
Classifier \cite{liu2021OPA} &  0.13 & 16.02 & 30.15 & 61.25 & 0.108 \\
\hline
Ours &  \textbf{0.11} & \textbf{27.04} & \textbf{50.70} & \textbf{89.65} & \textbf{0.052}\\
\hline

\hline
\end{tabular}}
\caption{Evaluation on scale prediction given ground-truth location on Pixabay. IOU is computed on the ground-truth location.}
\label{tab:scale}
\vspace{-0.1cm}
\end{table}

\subsection{Overfitting to Inpainting Artifact?}
Although we randomly apply inpainting on multiple regions in addition to the ground-truth location, the model may still overfit to the inpainting artifact by predicting only the inpainted regions. To understand whether the model is overfitting on the inpainting artifacts, we compare the inpainting mask and our prediction heatmap. We first select the highest score at each location of the predicted heatmap, resulting in a 2D heatmap. Then we binarize the 2D heatmap with a threshold so that the generated mask highlights  the same number of pixels as the inpainting mask. We compute the IOU between the prediction mask and the inpainting mask to see if there is a strong correlation. Since the location of the original object is also inpainted and trained as ground-truth, even a reasonable model is likely to predict high scores for the ground-truth inpainted regions, resulting in overlapping between the prediction mask and the inpainting mask. We observe that only a small portion ($\sim 14\%$) of predictions highlight a similar region ($IOU>0.5$) as the inpainting mask, which means the model is not always highlighting all the inpainted regions.  
We further show qualitative results in Fig. \ref{fig:artifact} to illustrate that the model is not overfitting to the inpainting artifacts, which is consistent with the observation in \cite{zhang2020learning}. 

    
    
    

\subsection{Generalization to Real-world Images}
To evaluate the generalization ability on real-world images, we collect a diverse set of pure background images with compatible foreground object images. The background images are selected by searching for keywords in Pixabay engine, including mountain road, city street, 
beach, island, indoor room, food, tree branch, etc. Some of them are very challenging, \eg inserting birds on branches as shown in Fig. \ref{fig:user}. In total, we select 24 background images with $3\sim6$ objects images for each, resulting in $100$ pairs for evaluation. Each of the $5$ methods generates $100$ results, leading to $500$ results for evaluation. We then shuffle the relative order of methods and ask users to rate all the $500$ samples in three levels: 0) ``Unsatisfactory'': The location and scale are clearly wrong. 1) ``Borderline'': The location and scale are OK but somewhat unrealistic. 2) The location and scale are clearly reasonable. Each sample is rated by at least five individuals and we report the average portion of each level.

\begin{table}[tbp]
\small
\vspace{-0.2cm}
\centering
\resizebox{\linewidth}{!}{
\begin{tabular}{l c c c }
\hline

\hline
Method &  Unsatisfactory $\downarrow$ & Borderline & Satisfactory $\uparrow$\\
\hline

\hline
Regression \cite{zhang2020learning}  & 46.8 & 17.4 & 35.8 \\
$\dagger$Retrieval \cite{zhu2022gala} & 45.4 & 22.6 & 32.0 \\  
Classifier \cite{liu2021OPA} & 72.0  & ~9.6 & 18.4 \\ 
PlaceNet \cite{zhang2020learning} & 69.0 & 12.6 & 18.4 \\ 
\hline
Ours & \textbf{42.8} & 17.8 & \textbf{39.4} \\ 
\hline

\hline
\end{tabular}}
\caption{Human evaluation on location and scale prediction for real-world images.}
\vspace{-0.3cm}
\label{tab:user}
\end{table}

\begin{figure*}[!htbp]
    \centering
    \includegraphics[width=.95\linewidth]{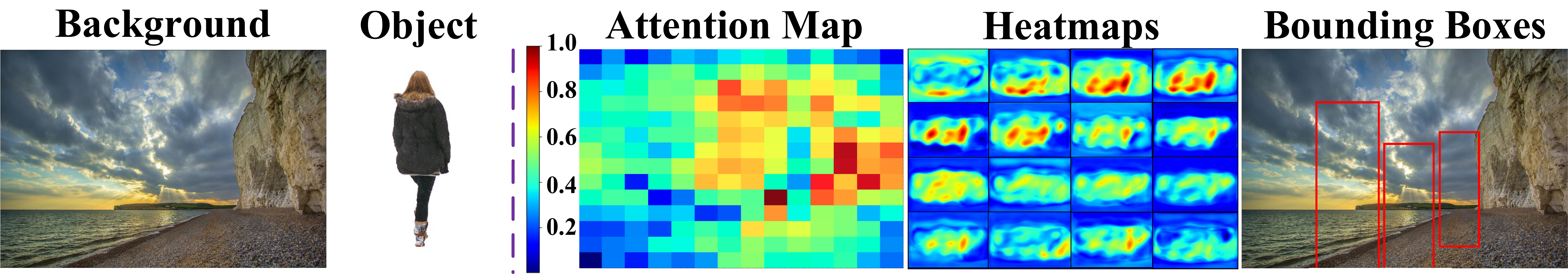}
    \vspace{-0.1cm}
    \caption{Example attention map between the object token and background local tokens, which highlights multiple candidate regions.}
    \label{fig:vis}
    \vspace{-0.3cm}
\end{figure*}

As shown in Table \ref{tab:user}, our method achieves a higher satisfactory (True Positive) rate and a lower unsatisfactory (False Positive) rate than all the other methods. Note that there is a domain gap between inpainted images and natural images, the performance of methods trained on inpainted images is somewhat degraded. The performance of our method could be further improved by finetuning with small-scale human annotation on real-world images. Qualitative results are provided in Fig. \ref{fig:user} to show that the proposed method performs better when local clue is critical for object placement.

\subsection{Ablation Study}
\noindent\textbf{Effect of Loss Functions.} We compare the proposed sparse contrastive loss with binary and Gaussian assignment loss (Sec. \ref{sec:loss}) and report the top-5 IOU (Sec. \ref{sec:metric}) on Pixabay dataset in Table \ref{tab:loss}. ``Binary Assignment'' and ``Gaussian Assignment'' both encourage a single-peak heatmap, but their performance is much lower than the proposed ``Sparse Contrastive'', indicating the margin-based contrastive design is necessary for this task. A multi-peak distribution is usually observed in the prediction of the proposed method, which aligns with the fact that there are multiple reasonable location/scale combinations to insert an object. 

\begin{table}[tbp]
\small
    \centering
    \begin{tabular}{l c c}
    \hline
    
    \hline
    Ablations & $IOU>0.5$ &  Mean $IOU$ \\
    \hline
    Binary Assignment & ~0.00 & 0.050  \\
    Gaussian Assignment & ~9.50 & 0.252 \\
    Sparse Contrastive & \textbf{74.74} & \textbf{0.620} \\
    \hline
    
    \hline
    \end{tabular}
    \caption{Effect of different loss functions.}
    \label{tab:loss}
    \vspace{-0.1cm}
\end{table}


\noindent\textbf{Effect of Feature Concatenation.} The are several ways to concatenate the local background features with the global object feature. ``Global Only'' simply concatenates the global features of both background and object images as the input of the upsampling decoder, which does not involve any local information. ``Local Concat.'' directly concatenates the global object feature to every local feature of the background image as the input of the upsampling decoder. In this case, every local feature has information from the object image, but there is no long-range correlation between different local features. ``Local Atten.'' combines the local background features and global object feature with a transformer, which enables the learning of correlation between any of these features with self-attention. In Table \ref{tab:arc}, ``Local Atten.'' achieves much better performance than other configurations in terms of top-5 IOU (Sec. \ref{sec:metric}) on Pixabay dataset, indicating the importance of learning correlation between local background features and the global object feature.

\noindent\textbf{Visualization.} In Fig. \ref{fig:vis}, we show an example of the learned attention map between the foreground object token and the background local tokens. We observe that it focuses on important local regions, which can not be learned by previous methods with only global features. This supports the motivation of our design to learn local clues for object placement. One complete view of $16$ heatmaps for all scales is also provided. Different locations could be recommended for different scales.
\begin{table}[tbp]
\small
    \centering
    \begin{tabular}{l c c}
    \hline
    
    \hline
    Ablations & $IOU>0.5$ &  Mean $IOU$ \\
    \hline
    Global Only & ~0.78 & 0.069 \\
    Local Concat. & 48.16 & 0.466 \\
    Local Atten. & \textbf{74.74} & \textbf{0.620}\\ 
      \hline
      
      \hline
    \end{tabular}
    \caption{Effect of feature concatenations.}
    \label{tab:arc}
    \vspace{-0.1cm}
\end{table}

    


\section{Discussion and Conclusion}
\label{sec:conclusion}
We propose a novel object placement method for real-world compositing, which generates dense evaluation on all pre-defined placements (location/scale of the bounding box) in a single network forward pass. It learns the correlation between object features and local background features using a transformer module so that local clues in background images can be leveraged to determine whether a particular placement is plausible. Experiments on both manually annotated dataset and large-scale inpainted dataset show significant improvements over previous state-of-the-art methods. It also generalizes well to challenging real-world cases. \\
\noindent\textbf{Limitation and Broader Impact.} One limitation is that our model only handles 2D image without considering 3D information, \eg lighting, shadow, and occlusion. We could add 3D modeling for both the object and background in the future to achieve more realistic object placement. Another limitation is that we rely on an off-the-shelf inpainting model for creating our training dataset, and there is a domain gap between inpainted images and natural ones. We can add semi-supervised fine-tuning on a small set of manually annotated images to improve the generalization ability. One possible negative societal impact is the bias of our system's performance on certain combinations of object categories and background types. More discussion is included in the supplementary material.

{\small
\bibliographystyle{ieee_fullname}
\bibliography{egbib}
}

\end{document}